\documentclass[12pt]{article}

\usepackage{makecell}
\usepackage{amssymb,amsmath,amsfonts,latexsym,graphicx}
\usepackage[numbers]{natbib}
\usepackage{hyperref} 
\hypersetup{
    colorlinks=true,
    urlcolor=blue,
    citecolor=black,
    pdftitle={Overleaf Example},
    pdfpagemode=FullScreen,
}

\begin{document}

\begin{center}
\Large \bf VISTA: Validation-Guided Integration of Spatial and Temporal Foundation Models with Anatomical Decoding for Rare-Pathology VCE Event Detection \rm






\vspace{1cm}


\large Bo-Cheng Qiu $\,^a$, \large Yu-Fan Lin $\,^a$, \large Yu-Zhe Pien $\,^b$, \large Chia-Ming Lee $\,^a$, \large Fu-En Yang $\,^c$, \large Yu-Chiang Frank Wang $\,^c$, \large Chih-Chung Hsu $\,^b,^*$

\vspace{0.5cm}

\normalsize


$^a$ National Cheng Kung University, Taiwan

$^b$ National Yang Ming Chiao Tung University, Taiwan

$^c$ NVIDIA, Taipei, Taiwan

\vspace{5mm}


Corresponding Author Email: {\tt \href{mailto:chihchung@nycu.edu.tw}{chihchung@nycu.edu.tw}} 

Team Name: {\tt ACVLab}

GitHub Repository Link: {\tt \href{https://github.com/Joe1007/ICPR-2026-RARE-VISION-Challenge}{https://github.com/Joe1007/ICPR-2026-RARE-VISION-Challenge}}

\vspace{1cm}

\end{center}

\abstract{
Capsule endoscopy event detection is challenging because diagnostically relevant findings are sparse, visually heterogeneous, and embedded in long, noisy video streams, while evaluation is performed at the event level rather than by frame accuracy alone. We therefore formulate the RARE-VISION task as a metric-aligned event detection problem instead of a purely frame-wise classification task. Our framework combines two complementary backbones, EndoFM-LV for local temporal context and DINOv3 ViT-L/16 for strong frame-level visual semantics, followed by a Diverse Head Ensemble, Validation-Guided Hierarchical Fusion, and Anatomy-Aware Temporal Event Decoding. The fusion stage uses validation-derived class-wise model weighting, backbone weighting, and probability calibration, while the decoding stage applies temporal smoothing, anatomical constraints, threshold refinement, and per-label event generation to produce stable event predictions. Validation ablations indicate that complementary backbones, validation-guided fusion, and anatomy-aware temporal decoding all contribute to event-level performance. On the official hidden test set, the proposed method achieved an overall temporal mAP@0.5 of 0.3530 and temporal mAP@0.95 of 0.3235.
}

\section{Motivation}\label{sec1}

Capsule endoscopy analysis is difficult because diagnostically relevant findings are sparse, visually heterogeneous, and embedded in long, noisy video streams \cite{lin2024divide}. In the RARE-VISION challenge, this is further complicated by severe class imbalance, strong appearance variation caused by blur, bubbles, illumination fluctuation, and viewpoint changes, as well as an event-level evaluation protocol based on temporal mAP rather than frame accuracy alone \cite{Lawniczak2025}. Consequently, even strong frame-wise predictions may still lead to weak final performance if they are temporally fragmented, poorly calibrated, or anatomically inconsistent.

Recent studies suggest that this task should not be treated as a purely frame-wise classification problem. Robust medical visual recognition depends not only on local discriminability but also on preserving temporal structure and cross-source consistency \cite{lee2025taming}. Long-sequence representation learning is particularly suitable for endoscopy videos, where longer temporal context improves downstream video understanding \cite{wang2025improving}. At the same time, large self-supervised vision models provide strong dense and global visual features without requiring task-specific backbone redesign \cite{simeoni2025dinov3}. These observations motivate our dual-backbone design: a temporal encoder for local video dynamics and a strong visual foundation model for subtle anatomical and pathological patterns.

The long-tailed multi-label setting further requires imbalance-aware learning and label-sensitive aggregation. Asymmetric loss improves robustness for rare tail classes in multi-label long-tailed settings \cite{park2023robust}, while weighted strategies informed by class frequency and disease co-occurrence improve rare disease discrimination in multi-label medical imaging \cite{lin2025weighted}. This motivates our use of heterogeneous loss functions across ensemble heads and validation-guided class-wise weighting during fusion.

Robust inference under unseen test conditions is also critical. Test-time adaptation and augmentation improve robustness under distribution shift \cite{dai2025free,zhou2025tegda,lin2024divide}, while temperature scaling improves confidence reliability when downstream decisions depend on threshold stability \cite{chanda2025evaluating}. In wireless capsule endoscopy, sequence modeling and temporal smoothing have also been shown to improve robustness when visual evidence is degraded \cite{nam2024deep}. These observations motivate our inference design, including validation-guided fusion, probability calibration, threshold tuning, and anatomy-aware temporal refinement.

Motivated by the above considerations, we formulate the task as a metric-aligned event detection pipeline rather than a purely frame-centric classifier. Specifically, we use complementary temporal and visual backbones for representation learning, diverse ensemble heads for imbalance-aware prediction diversity, validation-guided hierarchical fusion for label-sensitive aggregation and calibration, and anatomically informed temporal decoding to convert frame-wise probabilities into stable event predictions.

\section{Methods}\label{sec2}

\begin{figure}[htbp]
    \centering
    \includegraphics[width=0.95\linewidth]{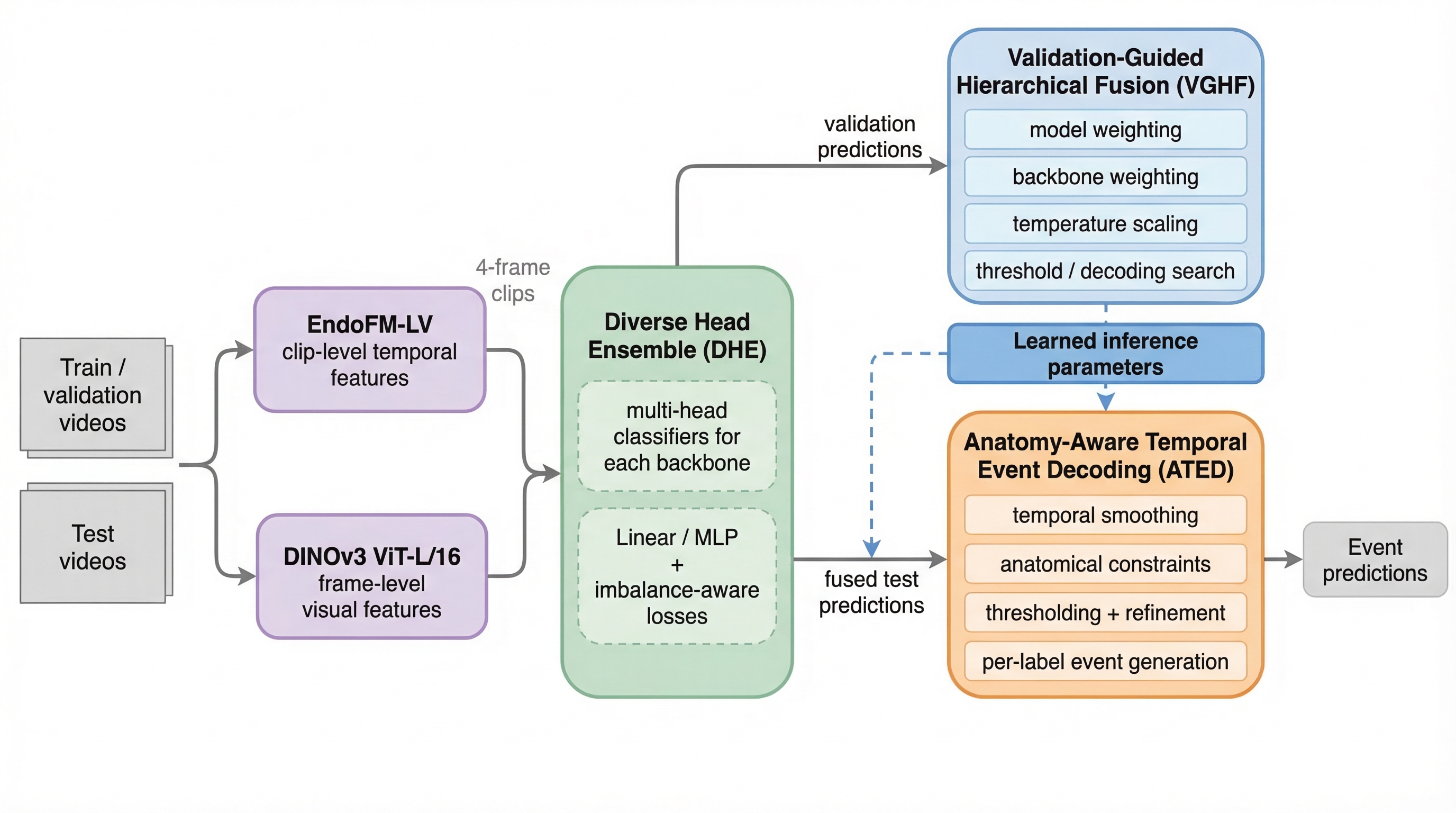}
    \caption{Overview of the developed pipeline.}
    \label{fig:enter-label}
\end{figure}

Figure~\ref{fig:enter-label} shows the overall pipeline. We extract complementary temporal and visual features using two backbones, train a Diverse Head Ensemble (DHE) on each feature stream, fuse predictions by Validation-Guided Hierarchical Fusion (VGHF), and convert fused frame-level probabilities into event predictions through Anatomy-Aware Temporal Event Decoding (ATED).

\subsection{Dual-Backbone Feature Extraction}

We use two complementary backbones on the same train/validation split. EndoFM-LV encodes short-range temporal context from a 4-frame clip with stride 2:
\begin{equation}
h_t^{(1)} = f_{\text{EndoFM}}(c_t),
\end{equation}
where $h_t^{(1)} \in \mathbb{R}^{768}$. DINOv3 ViT-L/16 extracts a frame-level visual representation
\begin{equation}
h_t^{(2)} = f_{\text{DINOv3}}(x_t),
\end{equation}
where $h_t^{(2)} \in \mathbb{R}^{1024}$. The two backbones provide complementary temporal and frame-level visual cues.

\subsection{Diverse Head Ensemble and Validation-Guided Hierarchical Fusion}

On top of each backbone, we train a \textbf{Diverse Head Ensemble (DHE)} with five lightweight classifiers using different architectures and loss functions, including linear or MLP heads optimized by BCE, focal loss, or asymmetric loss. For backbone $b$, head $m$, and class $c$, the predicted probability is denoted by $p_{t,b,m,c}$.

These outputs are fused by \textbf{Validation-Guided Hierarchical Fusion (VGHF)}. For each backbone, class-wise Average Precision (AP) on the validation set determines model-level weights:
\begin{equation}
\alpha_{b,m,c}=\frac{\mathrm{AP}_{b,m,c}}{\sum_{m'} \mathrm{AP}_{b,m',c}}, \qquad
\hat{p}_{t,b,c}=\sum_m \alpha_{b,m,c} p_{t,b,m,c}.
\end{equation}
The two backbone-level predictions are then fused using validation frame-level mAP:
\begin{equation}
\hat{p}_{t,c}=\sum_b \beta_b \hat{p}_{t,b,c}.
\end{equation}
We further calibrate the fused probabilities by temperature scaling,
\begin{equation}
\tilde{p}_{t,c}=\sigma\!\left(\frac{\mathrm{logit}(\hat{p}_{t,c})}{T}\right),
\end{equation}
where $T$ is selected by grid search on the validation set.

The same validation stage is also used to tune downstream inference parameters. Class-specific thresholds are initialized from precision--recall curves using F1 and refined by local search with temporal mAP as the target objective. Temporal smoothing and morphological refinement parameters are also selected on the validation set.

\subsection{Anatomy-Aware Temporal Event Decoding}

At test time, EndoFM-LV extracts clip-level features online and applies test-time augmentation by averaging predictions from original and horizontally flipped clips, while DINOv3 uses pre-extracted frame features. Predictions are fused using the validation-derived model weights, backbone weights, temperature, thresholds, and post-processing parameters.

The fused probability sequence is then processed by \textbf{Anatomy-Aware Temporal Event Decoding (ATED)}. First, class-specific temporal smoothing is applied independently to each video, with larger windows for anatomical classes and smaller windows for pathological classes. Second, we apply simple anatomical constraints during decoding. The five major anatomical regions are treated as mutually exclusive, so only the highest-probability region is retained at each frame. We also enforce the monotonic gastrointestinal transit order from \texttt{mouth} to \texttt{colon}, which suppresses anatomically implausible backward transitions. In addition, landmark labels are only retained when they appear within, or close to, their anatomically valid neighboring regions. Third, class-specific thresholds are applied to obtain binary sequences, followed by morphological opening and closing to suppress short false positives and reconnect fragmented events. If all anatomical regions are removed at a frame, the most likely region is reassigned to ensure full anatomical coverage.

Finally, event generation is performed independently for each label, where each contiguous positive segment is treated as one event. This per-label decoding is more faithful to temporal IoU-based evaluation than tuple-based multi-label segmentation because pathological events are not artificially split when anatomical labels change over time.

\subsection{Class-Imbalance Handling}

Class imbalance is handled at both training and inference time. During training, we use BCE with \texttt{pos\_weight}, focal loss, and asymmetric loss across different ensemble heads to improve learning for rare labels. During inference, imbalance is further mitigated by class-wise ensemble weighting and class-specific threshold optimization, allowing minority pathological classes to benefit from models and operating points with better recall.

\section{Results}\label{sec3}

\paragraph{Dataset and split.}
We follow the official ICPR 2026 RARE-VISION competition protocol based on the Galar dataset. The development cohort contains 80 labeled videos with frame-wise annotations and metadata, and is used for model training and validation. Following the competition setting, the official test set is strictly separated from the development data and consists of 3 previously unseen NaviCam examinations with withheld annotations for blind evaluation. As the competition protocol does not prescribe a fixed internal train/validation partition within the 80 development videos, validation was performed using our own split strategy on the released development cohort.

We report validation-set ablations using the official event-level metrics temporal mAP@0.5 and temporal mAP@0.95. Table~\ref{tab:ablation_main} evaluates the main design choices of the proposed framework, including complementary backbones, validation-guided fusion, and anatomy-aware temporal event decoding.

\begin{table*}[htbp]
\centering
\small
\setlength{\tabcolsep}{4pt}
\renewcommand{\arraystretch}{1.15}
\begin{tabular}{l l l c c}
\hline
\textbf{Backbone(s)} & \textbf{Fusion strategy} & \textbf{Event decoding strategy} & \textbf{mAP@0.5} & \textbf{mAP@0.95} \\
\hline
EndoFM-LV & Single backbone & \makecell[l]{Full ATED \\ + per-label events} & 0.3503 & 0.2845 \\
DINOv3 & Single backbone & \makecell[l]{Full ATED \\ + per-label events} & 0.4183 & 0.3309 \\
EndoFM + DINOv3 & \makecell[l]{Weighted model \\ + backbone fusion} & Per-label events only & 0.3002 & 0.2612 \\
EndoFM + DINOv3 & \makecell[l]{Weighted model \\ + backbone fusion} & \makecell[l]{Full ATED \\ + tuple-based events} & 0.3897 & 0.3333 \\
EndoFM + DINOv3 & \makecell[l]{Uniform model \\ + backbone fusion} & \makecell[l]{Full ATED \\ + per-label events} & 0.4520 & 0.3492 \\
EndoFM + DINOv3 & \makecell[l]{Weighted backbone / \\ uniform model} & \makecell[l]{Full ATED \\ + per-label events} & 0.4799 & 0.3388 \\
\textbf{EndoFM + DINOv3} & \makecell[l]{\textbf{Weighted model} \\ \textbf{+ backbone fusion}} & \makecell[l]{\textbf{Full ATED} \\ \textbf{+ per-label events}} & \textbf{0.4730} & \textbf{0.3658} \\
\hline
\end{tabular}
\caption{Validation-set ablation study of the proposed framework.}
\label{tab:ablation_main}
\end{table*}

The full method achieved the best overall balance across the two evaluation thresholds, reaching 0.4730 temporal mAP@0.5 and 0.3658 temporal mAP@0.95. This result supports three main conclusions.

First, the two backbones are complementary. DINOv3 alone outperformed EndoFM-LV alone (0.4183/0.3309 vs.\ 0.3503/0.2845), while the fused model further improved performance, indicating that EndoFM-LV contributes useful temporal information beyond a strong frame-level visual backbone.

Second, validation-guided fusion improves event detection over naive aggregation. Replacing weighted fusion with uniform model and backbone averaging reduced performance from 0.4730/0.3658 to 0.4520/0.3492. Using uniform model averaging with weighted backbone fusion slightly increased mAP@0.5 (0.4799 vs.\ 0.4730) but reduced mAP@0.95 (0.3388 vs.\ 0.3658), suggesting that class-wise model weighting is particularly beneficial for more precise event localization. We therefore retained fully weighted fusion in the final system.

Third, anatomy-aware temporal decoding is critical. Removing structured temporal refinement reduced performance from 0.4730/0.3658 to 0.3002/0.2612. Replacing per-label event construction with tuple-based segmentation also degraded performance to 0.3897/0.3333. These results show that both temporal refinement and per-label event generation are important for strong event-level prediction.

\paragraph{Official test-set results.}
We submitted the full method to the official competition scoring system, and the hidden-test results are shown in Table~\ref{tab:test_results}. The final submission achieved an overall temporal mAP@0.5 of 0.3530 and temporal mAP@0.95 of 0.3235, indicating that the main design choices remained effective on the hidden test set, although a noticeable validation-to-test gap remained.

\begin{table}[htbp]
\centering
\small
\setlength{\tabcolsep}{6pt}
\renewcommand{\arraystretch}{1.15}
\begin{tabular}{l c c}
\hline
\textbf{Video ID} & \textbf{mAP@0.5} & \textbf{mAP@0.95} \\
\hline
ukdd\_navi\_00051 & 0.4706 & 0.4412 \\
ukdd\_navi\_00068 & 0.2356 & 0.1765 \\
ukdd\_navi\_00076 & 0.3529 & 0.3529 \\
\hline
\textbf{Overall average} & \textbf{0.3530} & \textbf{0.3235} \\
\hline
\end{tabular}
\caption{Official hidden-test results obtained from the competition scoring system.}
\label{tab:test_results}
\end{table}

\section{Discussion}\label{sec4}

Overall, the results suggest that event-level capsule endoscopy detection depends not only on strong visual representations, but also on inference design. In particular, the validation study shows that validation-guided fusion and anatomy-aware temporal event decoding are important for converting noisy frame-wise predictions into stable event predictions. This indicates that temporal continuity and anatomically plausible transitions are central to the task.

At the same time, several limitations remain. Most fusion and decoding parameters are selected on a single validation split, which may limit generalization and partly explain the validation-to-test performance gap. In addition, DINOv3 is stronger than the temporal branch when used alone, suggesting that temporal modeling in the current framework is still relatively weak. Future work should therefore focus on more robust parameter selection, stronger temporal representation learning, and improved robustness to challenging visual conditions and cross-video domain shift.

\section{Summary}\label{sec5}


Team ACVLab utilized a multi-backbone event detection framework combining EndoFM-LV and DINOv3 ViT-L/16. A combination of BCE with \texttt{pos\_weight}, focal loss, asymmetric loss, class-wise ensemble weighting, and class-specific threshold optimization was used to handle class imbalance in the dataset. The results show that structured temporal decoding and per-label event construction are critical for strong event-level performance in capsule endoscopy videos. The team achieved an overall mAP@0.5 of 0.3530, and overall mAP@0.95 of 0.3235.

\section{Acknowledgments}\label{sec6}

As participants in the ICPR 2026 RARE-VISION Competition, we fully comply with the competition's rules as outlined in \cite{Lawniczak2025}. Our AI model development is based exclusively on the datasets in the competition. The mAP values are reported using the test dataset and sanity checker released in the competition.

\bibliographystyle{unsrtnat}
\bibliography{sample}

\end{document}